\documentclass[letterpaper, 10 pt, conference]{ieeeconf}  
\IEEEoverridecommandlockouts                          
\usepackage{amsmath,amssymb,amsfonts}
\usepackage{dsfont}
\usepackage{graphicx}
\usepackage{textcomp}
\usepackage{xcolor}
\usepackage{float}
\usepackage{booktabs}
\usepackage{multirow}
\usepackage{todonotes}
\usepackage{multirow}
\usepackage{graphicx}
\usepackage{algorithm}
\usepackage{amsmath}  
\usepackage[colorlinks]{hyperref}
\hypersetup{colorlinks, linkcolor=red}

\def\BibTeX{{\rm B\kern-.05em{\sc i\kern-.025em b}\kern-.08em
    T\kern-.1667em\lower.7ex\hbox{E}\kern-.125emX}}
\title{\LARGE \bf
Robust Change Detection Based on Neural Descriptor Fields}

\author{Jiahui Fu, Yilun Du, Kurran Singh, Joshua B. Tenenbaum, and John J. Leonard
\thanks{All authors are with the MIT Computer Science and Artificial Intelligence Laboratory, Cambridge, MA 02139, USA.
        {\tt\{jiahuifu,yilundu,singhk,jbt,jleonard\}@mit.edu}}
\thanks{This work was supported by ONR MURI grant N00014-19-1-2571 and ONR grant N00014-18-1-2832.}}

\usepackage[backend=bibtex,
            hyperref=false,
            url=false,
            isbn=false,
            doi=false,
            backref=false,
            style=numeric-comp,
            sorting=none,
            sortcites,
            block=none]{biblatex}
            
\renewcommand{\bibfont}{\small}
\newcommand{\sethree}{$\text{SE}(3)$}
\newcommand{\sothree}{$\text{SO}(3)$}
\begin{document}

\maketitle
\thispagestyle{empty}
\pagestyle{empty}

\begin{abstract}
The ability to reason about changes in the environment is crucial for robots operating over extended periods of time. Agents are expected to capture changes \emph{during operation} so that actions can be followed to ensure a smooth progression of the working session. However, varying viewing angles and accumulated localization errors make it easy for robots to falsely detect changes in the surrounding world due to low observation overlap and drifted object associations. In this paper, based on the recently proposed category-level Neural Descriptor Fields (NDFs), we develop an object-level online change detection approach that is robust to partially overlapping observations and  noisy localization results. Utilizing the shape completion capability and {\sethree}-equivariance of NDFs, we represent objects with compact shape codes encoding  \emph{full} object shapes from partial observations. The objects are then organized in a spatial tree structure based on object centers recovered from NDFs for fast queries of object neighborhoods. By associating objects via shape code similarity and comparing local object-neighbor spatial layout, our proposed approach demonstrates robustness to low observation overlap and localization noises. We conduct experiments on both synthetic and real-world sequences and achieve improved change detection results compared to multiple baseline methods. Project webpage: \href{https://yilundu.github.io/ndf_change}{http://yilundu.github.io/ndf\_change}
\end{abstract}

\section{Introduction}
The ability to perform robust long-term operations is critical for many robotics applications such as room scanning and household cleaning. As these tasks usually involve frequent visits to the same environment over extended periods of time, during which the environment may experience changes, robots are expected to understand these newly-emerged scene differences, e.g., the introduction and removal of a coffee mug, as they may impact the proper subsequent actions to be taken \emph{during} operation. 

An intuitive way to conduct change detection is to perform scene differencing between current observations and scenes reconstructed from previous sessions. Using various scene representations such as point clouds and Truncated Signed Distance Fields (TSDF),  previous works detect changed areas in the environment through global point-wise or voxel-wise differencing on two scenes reconstructed and pre-aligned \emph{post hoc} from sequential data~\cite{finman2013toward,ambrucs2014meta,fehr2017tsdf}. These methods, which treat the environment as an unordered collection of points or voxels, demand high inter-session viewpoint overlap and are susceptible to noisy sensor data and localization errors. Considering the fact that changes take place at the object level, recent works~\cite{langer2020robust,schmid2022panoptic} explore the use of semantic consistency for local object-level verification on top of the common global point- or voxel-wise scene comparison scheme. Having demonstrated improved robustness to localization errors, they are nevertheless prone to failure with noisy reconstruction input and little observation overlap, both of which are frequently encountered during online change detection tasks.
\begin{figure}
\vspace{-5pt}
    \centering
    \includegraphics[width=0.85\linewidth]{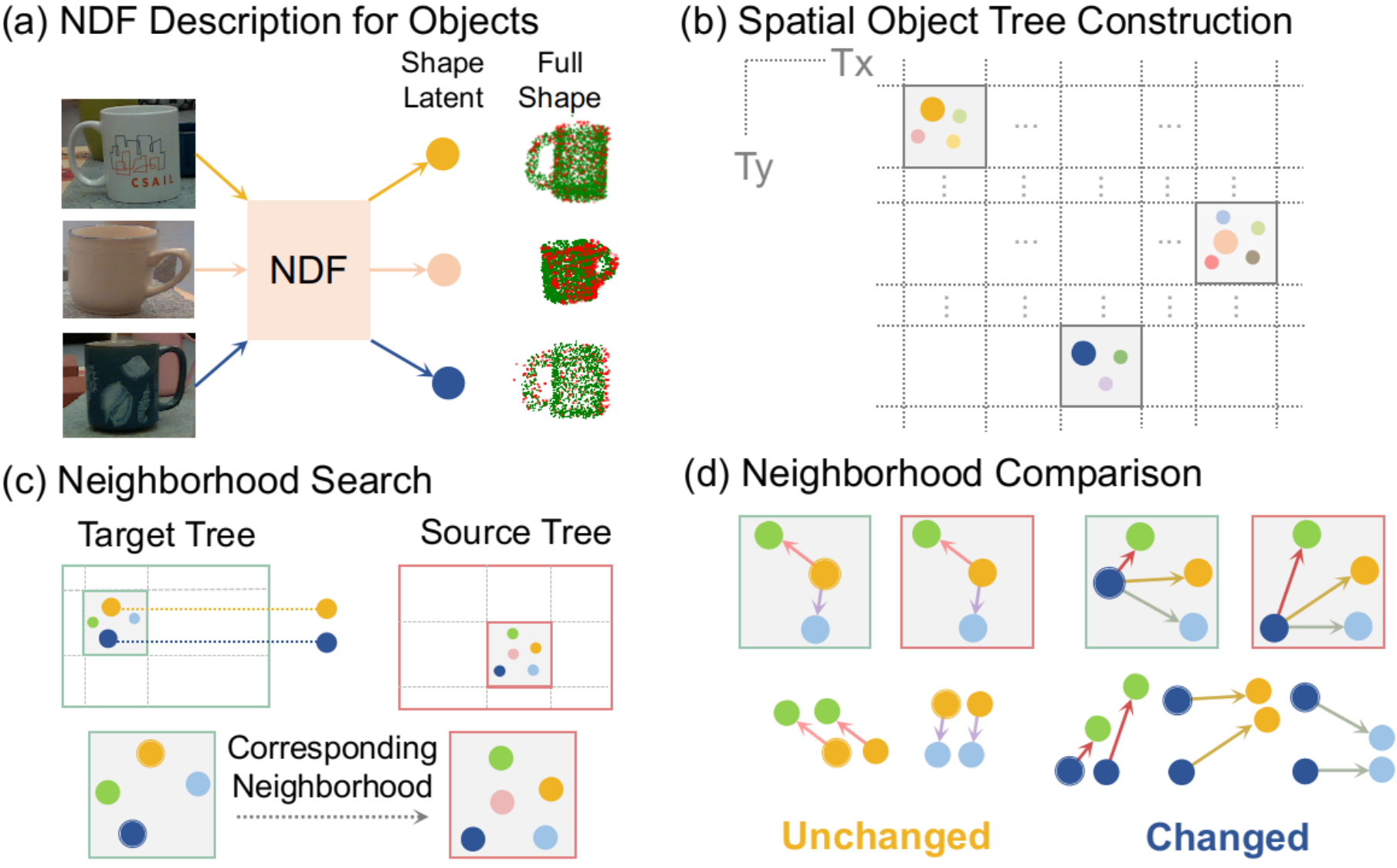}
    \caption{Approach overview. Given a sequence of observations as the source and the target, during streaming of the target sequence, the system takes in partial object point clouds from depth sensors and outputs changed objects on-the-fly in the target w.r.t the source. (a) Given each partial object point cloud, Neural Descriptor Fields (NDFs) represent the object as a shape code encoding the full object shape and recover the object center from full shape reconstruction. (b) Based on recovered object centers, observed objects are organized in a spatial object tree, which consists of two coordinate interval trees $(T_x, T_y)$ and allows for fast query of neighboring objects. (c)-(d): Corresponding neighborhoods of the current object are found from the source and the target object tree using object locations, where objects of similar shape codes are matched. For each matched object pair, object graphs of the neighborhood are constructed and compared to determine if the local object layouts are different, which implies a changed object. }
    \label{pipeline}
    \vspace{-20pt}
\end{figure}

It is therefore highly desirable to seek approaches that yield consistent representations of the object surmounting viewing angle limitations, where we then adopt the recently proposed Neural Descriptor Fields (NDFs)~\cite{9812146}. NDFs have been developed as a category-level, {\sethree}-equivariant object representation for object manipulation tasks. By encoding object as a continuous {\sethree}-equivariant function, NDFs are able to reconstruct the full object shape through a compact shape encoding and further recover object centers (translation). Considering that NDFs are formulated to only ensure identical shape codes given rotated point clouds that are otherwise identical, we augment its shape completion capability by enforcing a new shape similarity loss with partial object observations as inputs, making it encode consistent full object shape and positions across different viewing angles.

Following the object-level interpretation of the world via NDFs-derived object representations, in this paper, we propose an object change detection approach (see Fig.~\ref{pipeline}) for mobile robots, targeting the most common online operating scenarios with high viewing angle variation, potential localization errors, and no pre-alignment tools available. Based on the {\sethree}-equivariance property and category-level shape completion ability of NDFs, we represent the object with a compact \emph{full} shape code obtained from partial observations, and organize the objects in a spatial tree structure in terms  of object centers recovered from NDFs for fast query of object neighborhoods. By associating objects through shape code similarity and decomposing the global differencing scheme into local object-neighbor layout comparison, our approach shows improved robustness to little observation overlap and  localization errors.
 Our main contributions are as follows:
\begin{enumerate}
    \item First, we explore the use of NDFs for category-generalizable shape consistent object representation across different viewing angles.
    \item Next, we propose an online change detection approach based on NDF-derived object representations and local object layout comparison.
    \item Finally, we demonstrate the effectiveness of the proposed approach on both synthetic and real-world testing sequences featuring novel same-category object instances not seen during training time.
\end{enumerate}

\section{Related Work}
\subsection{Change Detection}
Previous works conduct change detection based on inputs in various 2D and 3D environment representations. Given 2D image inputs, Derner et al.~\cite{DERNER2021103676} build a visual database and compare classical 2D feature descriptors extracted from greyscale images for query and reference feature matching. For works with 3D representations, Finman et al.~\cite{finman2013toward} discover new objects as changed parts of multiple depth point clouds while Herbst et al. detect changes through movement of surfaces~\cite{5980542}. Ambrus et al.~\cite{ambrucs2014meta} compute a meta-room static reference map based on point clouds, and discover new objects from  changes via spatial clustering.  Fehr et al.~\cite{fehr2017tsdf} propose a multi-layer TSDF grid structure and perform volumeric differencing for object discovery and class recognition. Langer et al.~\cite{langer2020robust} combine semantic information and supporting plane information to discover objects newly introduced to the scene. Schmid et al.~\cite{schmid2022panoptic} propose a multi-TSDF panoptic mapping approach and conduct online change detection based on TSDF value comparison.

Most previous works focus on point-wise or voxel-wise differencing on pre-aligned reconstructions in an offline fashion, demanding high observation overlap and decent localization results. They therefore  do  not  always  satisfy  the need for online change detection under various viewing angles, as commonly encountered during mobile robot operation.

\subsection{Neural Implicit Representations}
 Neural implicit representations have been proposed as a continuous, differentiable, and parameterized representation of 3D geometry ~\cite{Niemeyer2019ICCV,park2019deepsdf,peng2020convolutional}, appearance \cite{sitzmann2019srns,mildenhall2020nerf}, and tactile properties~\cite{gao2021objectfolder} of both objects and scenes. Neural implicit representations represent shapes as continuous functions,  enabling the principled incorporation of symmetries, such as \sethree{} equivariance~\cite{Deng_2021_ICCV,9812146}, as well as the construction of latent spaces that encode class information as well as 3D correspondences~\cite{deng2021deformed}. Owing to their continuous parameterized nature, several works have applied~\cite{9812146, sucar2020nodeslam} them to robotics tasks as intermediate object representations directly inferable from raw perception. Simenov et al.~\cite{9812146} show that symmetries incorporated from such a representation can generalize demonstrations of objects to novel poses, while~\cite{sucar2020nodeslam,lionar2021neuralblox,huang2021di} demonstrate their ability to be integrated with online robotic mapping tasks. 
 
 In this paper, we explore how the \sethree-equivariance of NDFs~\cite{9812146} enables them to represent objects from partial observations and thus allows for robust online change detection with disparate viewing angles and localization noises.

\section{Category-level Object Representation for Partial Observations}
We base our object representation on the recently proposed Neural Descriptor Fields (NDFs)~\cite{9812146}, a function $f_{\theta}$ that encodes both object shape and pose through a category-level {\sethree}-equivariant latent representation. 
\subsection{Neural Descriptor Fields}
NDFs consist of an encoder function $f_{\theta}({\bf{P}) = z}$, which maps a partial object point cloud $\bf P$ into a global latent code $\bf z$, and a decoder function, $\Phi(\bf{x}, f_{\theta}(\bf{P}))$, which maps an input point $\bf{x}$ to its predicted occupancy value:
\begin{equation}
\begin{aligned}
   & f_{\theta}({\bf{P) = z}} :  \mathbb{R}^{n\times 3}\rightarrow\mathbb{R}^{k\times 3} \\
   & \Phi({\bf{x}, f_{\theta}({\bf{P}}))} = \Phi {\bf(\bf{x}, {\bf{z}}}) : \mathbb{R}^{3}\times \mathbb{R}^{k\times 3} \rightarrow [0,1].
\end{aligned}
\end{equation}
The encoder function $f_{\theta}(\cdot)$ is constructed such that by zero-centering the input $\bf P$, the inferred global latent code $\bf z$ is represented as a vector of points that is equivariant with respect to {\sothree} rotations of the input point cloud $\bf P$. This means that if a point cloud is rotated by $R$, the inferred latent will be equivalently rotated by $R$, as ensured via the Vector Neuron encoder layers~\cite{Deng_2021_ICCV}. By subtracting from $\bf x$ the point cloud center of $\bf P$ as the translation, we then make $f_{\theta}(\cdot)$ an {\sethree}-equivariant shape occupancy predictor for different points $\bf x$ on and off $\bf P$.

NDFs are trained with partial object point clouds recovered from semantic-segmented RGB-D images and corresponding 3D occupancy voxelgrids of objects' complete geometry. During training, the full model $[f_{\theta},\Phi]$ is trained to predict the complete 3D occupancy of an object using the  standard cross-entropy classification loss $L_{occ}$:
\begin{equation}
    L_{occ} = \mathcal L(\Phi({\bf{p}}, f_{\theta}({\bf{P}}),v),
\end{equation}
where $\bf{p}$ is a point sampled from the object occupancy grid and $v$ is the ground truth occupancy value at point $\bf{p}$.  

By feeding $\Phi(\cdot,\cdot)$ with a query point cloud $\mathcal{X}$, which is obtained via uniform sampling within a large bounding box centered around $\bf P$, the full shape point cloud $\mathcal{S}$ of the object can be reconstructed  in terms of the predicted occupancy values:
\begin{equation} \label{full}
    \mathcal{S} = \{{\bf x}| \Phi({\bf x},f_{\theta}({\bf x}\vert {\bf{P}}))>v_0, {\bf x}\in \mathcal{X}\},
\end{equation}
where $v_0$ is the occupancy threshold to mark a point location as occupied.

\subsection{Shape Consistency } Thanks to the {\sothree}-equivariance of ${\bf z}$, a shape code invariant of view direction, ${\bf{s}}\in \mathbb{R}^k$, can then be extracted from its rotation invariant portion:
\begin{equation}{\label{shapecode}}
    s_i = \vert\vert {\bf{z}}_i \vert\vert_2, i = 1,2,...,k.
\end{equation}

Ideally, the shape code,  serving as a compact representation of the full object shape, should be consistently close among partial observations of the same shape from various viewing perspectives, while discriminatively far apart across observations of different shapes. While NDFs enable identical shape codes given identical point clouds that are rotated, shape consistency across partial inputs of the same shape is not inherently guaranteed.

To enforce $\bf{s}$ to be a discriminative yet consistent representation for shapes seen from various viewing angles, as commonly seen during mobile robot operation, we further formulate a shape similarity loss, $L_{shape}$, fashioned after the idea of the triplet loss with [\emph{anchor, positives, negatives}] for person re-identification tasks~\cite{hermans2017defense}. Taking a partial point cloud from object shape $i$ as the anchor ${\bf{A}}_i$, we assign it with a positive sample ${\bf{P}}_i$ as the point cloud obtained from another perspective of the same object, and any other partial observations of a different object as the negative sample ${\bf{N}}_i$. With the distance metric $D(\cdot,\cdot)$ chosen as the cosine similarity, $L_{shape}({\bf{A}}_i,{\bf{P}}_i,{\bf{N}}_i)$ then tends to pull the shape codes of ${\bf{A}}_i$ and ${\bf{P}}_i$ closer, while pushing those of ${\bf{A}}_i$ and ${\bf{N}}_i$ further apart:
\begin{equation}\label{shape_sim}
    L_{shape} = -D({\bf{s}}_{{\bf{A}}_i},{\bf{s}}_{{\bf{P}}_i})+D({\bf{s}}_{{\bf{A}}_i},{\bf{s}}_{{\bf{N}}_i}).
\end{equation}

Moreover, to ensure that  $L_{shape}$ always finds the more informative $(\bf{A},\bf{P},\bf{N})$ triplet, we adopt the \emph{batch hard} way for triplet forming~\cite{hermans2017defense}, i.e., using the most dissimilar $(\bf{A},\bf{P})$ and the most similar $(\bf{A},\bf{N})$ within each batch to guide training. 

We hence populate each training batch $B$ with randomly generated observations of $({o_A},o_{P})$ pairs of different (but likely repetitive) object shapes, which ensures that at least two observations exist for each object within the batch. Every sample within the batch is then treated as an anchor, and paired batch-wise for the most dissimilar positive and the most similar negative samples. Hence the final batch-hard shape similarity loss is formulated as:
\begin{equation}
\begin{aligned}
        L_{b\_shape} = \frac{1}{\vert B \vert} \sum_{i=1}^{N}\sum_{j=1}^{N_i}(&-\min_{k\in[1,N_i]}D(o_{ij},o_{ik})\\
        &+\max_{m\neq i}D(o_{ij},o_{mn})),
\end{aligned}
\end{equation}
where $N$ is the number of objects whose observations are included in the current batch, $N_i$ the number of samples of object $i$ and $o_{ij}$ the $j$th observation of object $i$ within batch.

The final training objective is therefore formulated as the weighted combination of the cross entropy loss and the shape similarity loss as:
\begin{equation}
    L = L_{occ}+ \alpha L_{b\_shape},
\end{equation}
where $\alpha$ is the weighting coefficient set as $\alpha=$ 0.01.
\begin{figure}
    \centering
    \includegraphics[scale=0.15]{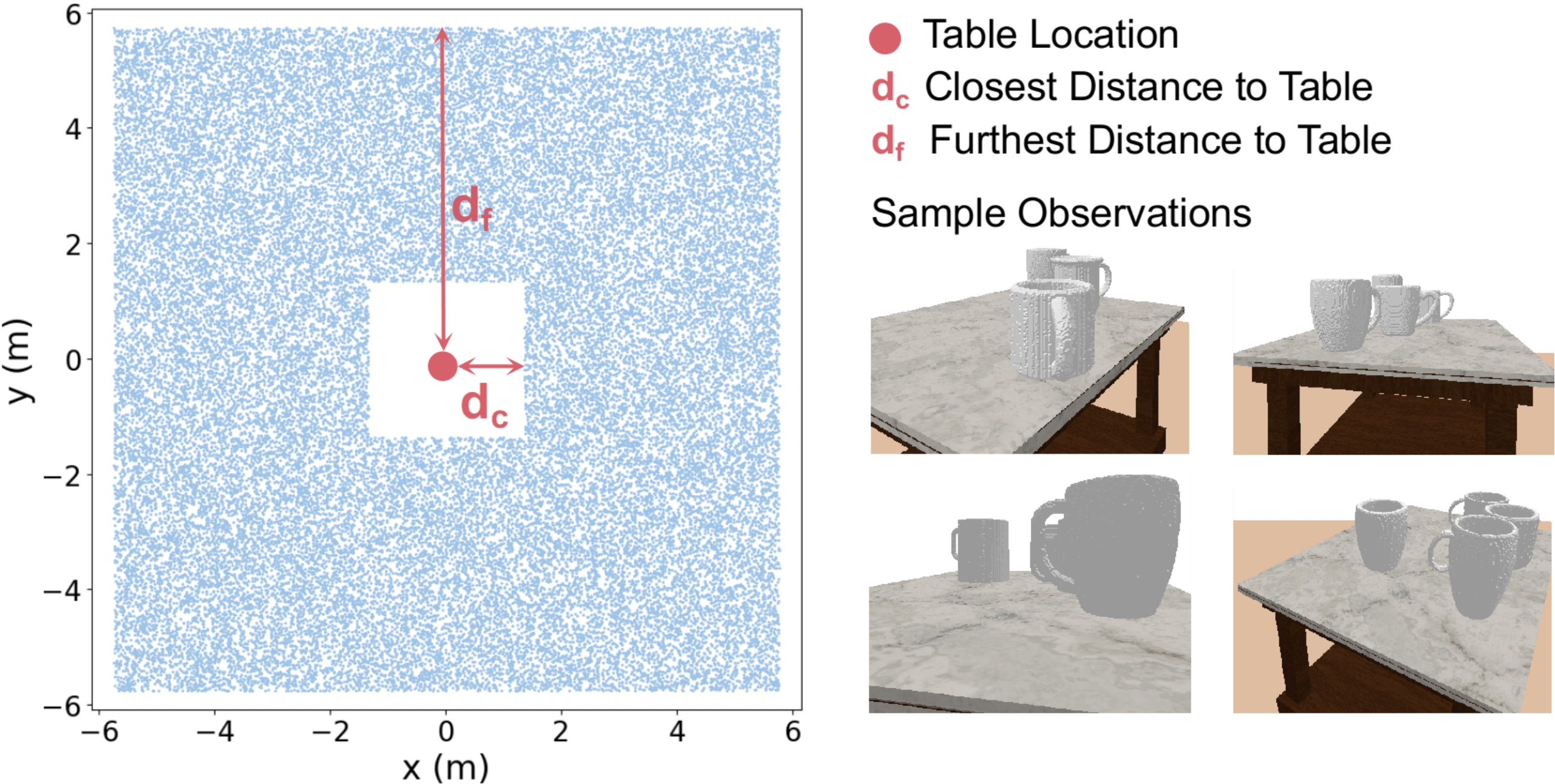}
    \vspace{-10pt}
    \caption{Training sample generation. Snapshots of the rendered scenes are presented on the right. Along with a height uniformly varying within 0.1 m from the tabletop, 2D camera locations (blue dots) are drawn uniformly from the hollow rectangular regions centered around the table. The width of the region is determined by the closest and furthest camera-to-table distance $d_c$ and $d_f$. $2-4$ geometric object models are randomly selected and placed on the table to simulate potential occlusions. }
    \label{sim}
    \vspace{-15pt}
\end{figure}

\subsection{Training in Simulation }\label{gen_sim} 
To overcome the data availability issue, we train our category-level shape-consistent NDFs fully in simulation using depth images rendered with Pybullet~\cite{coumans2016pybullet}. To consider the commonly encountered variation in viewing angles, observation distances, and occlusions during operation, for each training sample, we place a randomly-posed object on the table, along with 2-4 randomly drawn objects around it to create occlusion. The camera locations are sampled in a hollow rectangular region with the table at the center, thus accounting for both nearby and longer distance observations seen in real-world operations (see Fig.~\ref{sim}).

\subsection{Object Representation } We hence base our object representation  on the shape code in (\ref{shapecode}) and the object center (translation) $\bf{t}$ recovered from NDFs as $(\bf{s},t)$, where $\bf{t}$ is simply the predicted center of the reconstructed full point cloud $\mathcal{S}$ in (\ref{full}):
\begin{equation}
    {\bf{t}} = \frac{1}{\vert \mathcal{S} \vert} \sum_{{\bf{x}}\in \mathcal{S}} {\bf{x}}.
\end{equation}
By converting $\bf{t}$ into the world frame with the given camera pose, the object is characterized by $\bf{s}$ for its full shape and $\bf{t}$ for its global location. With the shape completion ability of NDFs, the current object representation is expected to provide a compact as well as robust way to distinguish object instances, even under varying viewing angles during robot motion.

\section{NDF-based Object Change Detection}\label{change_detection}
Inspired by the idea that local relative comparison is less sensitive to global localization drift than its global counterpart~\cite{langer2020robust}, we organize the observed object instances in a spatial object tree based on the predicted object centers, allowing for the convenient query of object neighborhoods. By comparing local neighboring object layout,  we improve the method's robustness to localization drift by avoiding direct comparison on the absolute values of localization results.
\subsection{Spatial Object Tree Construction}
To enable the online execution of our approach, we construct and maintain a spatial object tree in the world coordinate to organize the incoming data stream.

Built on top of the coordinate interval tree proposed in~\cite{zhang2020fusion}, our object tree comprises two translation interval trees,  $T=(T_x,T_y)$, in the $x$ and $y$ direction, respectively, which are initialized and updated simultaneously in the same fashion each time a new object measurement $(\bf s,t)$ arrives.
Considering that adjacent objects are located mostly on the same plane, for our case, it suffices to maintain trees only in the $x$ and $y$ direction for neighborhood query (while an extension to the $z$ direction should be straightforward).

\textbf{Tree Construction. } Consider $T_x$ for instance.  $T_x$ divides the $x-y$ plane into several equi-distant interval slices in the $x$ direction (Fig.~\ref{pipeline}(b)), where each instantiated interval $[x_{min},x_{max}]$ of fixed length $l = x_{max}-x_{min}$ is represented by a node $n$.  Each node stores a set of object instances whose translation component ${\bf t}$ follows ${t}_x \in [x_{min},x_{max}]$. The $x_{min}$ and $x_{max}$ are determined by the first ${t}_x$ that initializes this new node as $[{t}_x-l/2,{ t}_x+l/2]$. The new node is then inserted in the tree such that the upper interval limit of a left child $n_l$ is always smaller than the lower interval limit of its parent node $n_p$, i.e., $x_{l,max}<x_{p,min}$ and similarly for the right child $n_r$, we have $x_{r,min}>x_{p,max}$.

\textbf{Intra-tree Update. } We associate the incoming object measurement $\bf m=(\bf s,t)$ with existing object instances in the tree through spatial proximity  and the shape cosine similarity used in (\ref{shape_sim}). We first traverse through $T_x$ and $T_y$ to locate the corresponding coordinate interval nodes $n_x$ and $n_y$ that ${t}_x$ and ${ t}_y$ land in, respectively. By finding the intersection of the objects within $n_x$ and $n_y$, we obtain the neighborhood $\mathcal{N}$ that $\bf m$ should be adjacent to as $\mathcal{N}=\{o\vert o\in n_x \wedge o\in n_y\}$.
The measurement is then successfully associated with an object instance ${o}_0=({\bf s}_0,{\bf t}_0)$ in the neighborhood when:
\begin{equation}\label{association}
\begin{aligned}
      \vert \vert {\bf t}_0-{\bf t} \vert \vert <\delta_d \wedge D({\bf s}_0,{\bf s})>\delta_s, 
\end{aligned}
\end{equation}
where $\delta_d$ and $\delta_s$ are the maximum distance threshold and minimum shape similarity threshold for valid association. We then replace the old shape code and the recovered object center of the object with $\bf m$ as an update if the average occupancy value of the reconstructed full shape point cloud for $\bf m$ (calculated in (\ref{full})) is higher than that of $o_0$. 

\subsection{NDF-based Change Detection}
Given two sequences as the source and the target, we construct the source object tree $T_{s}$ in advance and perform object change detection during construction of the target tree $T_t$.

\textbf{Sequence Registration. } We wish to apply our approach in an online mobile operating scenario where post-alignment tools or motion capture systems are not always readily available. Therefore, we first conduct a rough alignment between the target and the source sequence at the beginning of the target traverse, in the hope that target objects can correctly locate the corresponding neighborhood patch in the source so as to enable \emph{valid} neighboring object comparison.

With all the shape codes of the source objects easily obtainable from $T_s$, after intra-tree update for $T_t$, we determine  each corresponding source object ${o}'$ for the current target object  ${o}$ as the one sharing the highest shape code similarity among all the objects (if exist) whose shape similarity with $o$ is above $\delta_s$ (similar to (\ref{association}) amid intra-tree measurement-object association).

After $N$ pairs of object correspondences ${(o',o)}$ have been accumulated, we apply Single Value Decomposition (SVD) with RANSAC onto the $N$  object center pairs $({\bf t'}_i,{\bf t}_i)$ and obtain the relative transform $T_{rel}$ between the two sequences.  This considers the potential inclusion of changed object instances, while assuming that they only take up a small portion of the environment to allow for alignment when the target sequence starts. Here, we set $N=6$ to account for transform accuracy versus timing to start change detection. 
\begin{figure}
    \centering
    \includegraphics[width=0.9\linewidth]{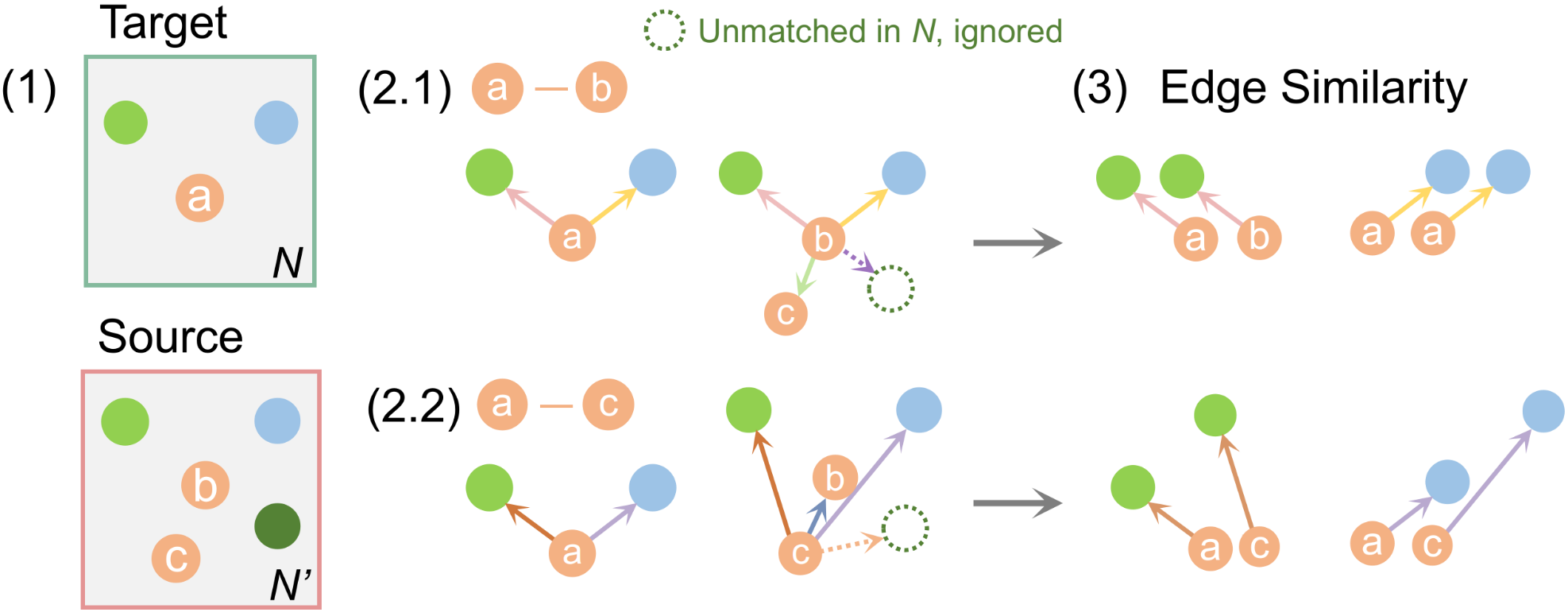}
    \caption{Neighborhood comparison as object graph matching. (1) Shape $a$ in the target neighborhood $\mathcal{N}$ forms two shape-similar pairs $(a,b)$ and $(a,c)$ from the corresponding source neighborhood $\mathcal{N}'$ (dots colored in orange). (2) Object graphs centered at $a$, $b$, and $c$ are constructed respectively, demonstrating local inter-object layout. Graphs of each pair are compared as shown in (2.1) for $(a,b)$ and  (2.2) for $(a,c)$. Each object graph consists of vertices as all the objects in the neighborhood (colored dots) and directed edges pointing from the object to the neighbor with length as  the object center difference  (colored arrows, e.g., ${\bf e}_{cb}={\bf t}_b-{\bf t}_c$). (3): Graph (neighborhood layout) comparison is conducted edge-wise, where edge similarity between vertex-matched edges (shown in arrows of the same color) is measured by the Euclidean distance between the two edge vectors.}
    \label{graph}
    \vspace{-15pt}
\end{figure}

\textbf{Change Detection. }Rather than examining object-wise correspondences through spatial proximity and shape similarity (as in (\ref{association})), we draw support from neighboring objects and  detect changes through the relative spatial layout consistency of the object within its target neighborhood and that of its corresponding neighborhood (if exists) in the source.

When an object measurement $\bf m=(s,t)$ arrives, we update the $T_t$ by finding/instantiating the associated object instance $O$. Considering the online operation setting, change detection is then performed on $O=({\bf s}_0,{\bf t}_0)$ that has \emph{not} been marked as changed before. 

Given $O=({\bf s}_0,{\bf t}_0)$ along with its neighborhood $\mathcal{N}_0$ in the target, we find the projected location in $T_s$ with $T_{rel}$ and query the corresponding source neighborhood $\mathcal{N}'=\{o'_i=({\bf s}_i,{\bf t}_i)\}$. Potential object matches are found based on shape code similarity between ${\bf s}_0$ and each ${\bf s}_i$ in $\mathcal{N}'$. 

The most straightforward change cases are that either $\mathcal{N}'$ is empty or no matched $o_i$ is found in the source neighborhood, as both indicate newly added objects to new locations or else new shapes.

Then with object matches found within $\mathcal{N}'$, we conduct local neighborhood comparison for each matched pair $(O,o'_i)$ (see Fig.~\ref{graph} for a concrete example). This also serves as a validation step for rejecting false positives in dealing with noise-corrupted localization. When object position changes at a scale similar to cross-session localization errors, it can be hard to determine if the absolute object center difference between $(O,o'_i)$ is brought by actual changes (e.g., object moves within the plane) or merely localization/projection drift, while the latter can be validated through the relatively more stable inter-object spatial layout.

For each of the matched pairs, neighborhood comparison is accomplished via object graph matching.  
\begin{figure}
    \centering
    \includegraphics[width=0.90\linewidth]{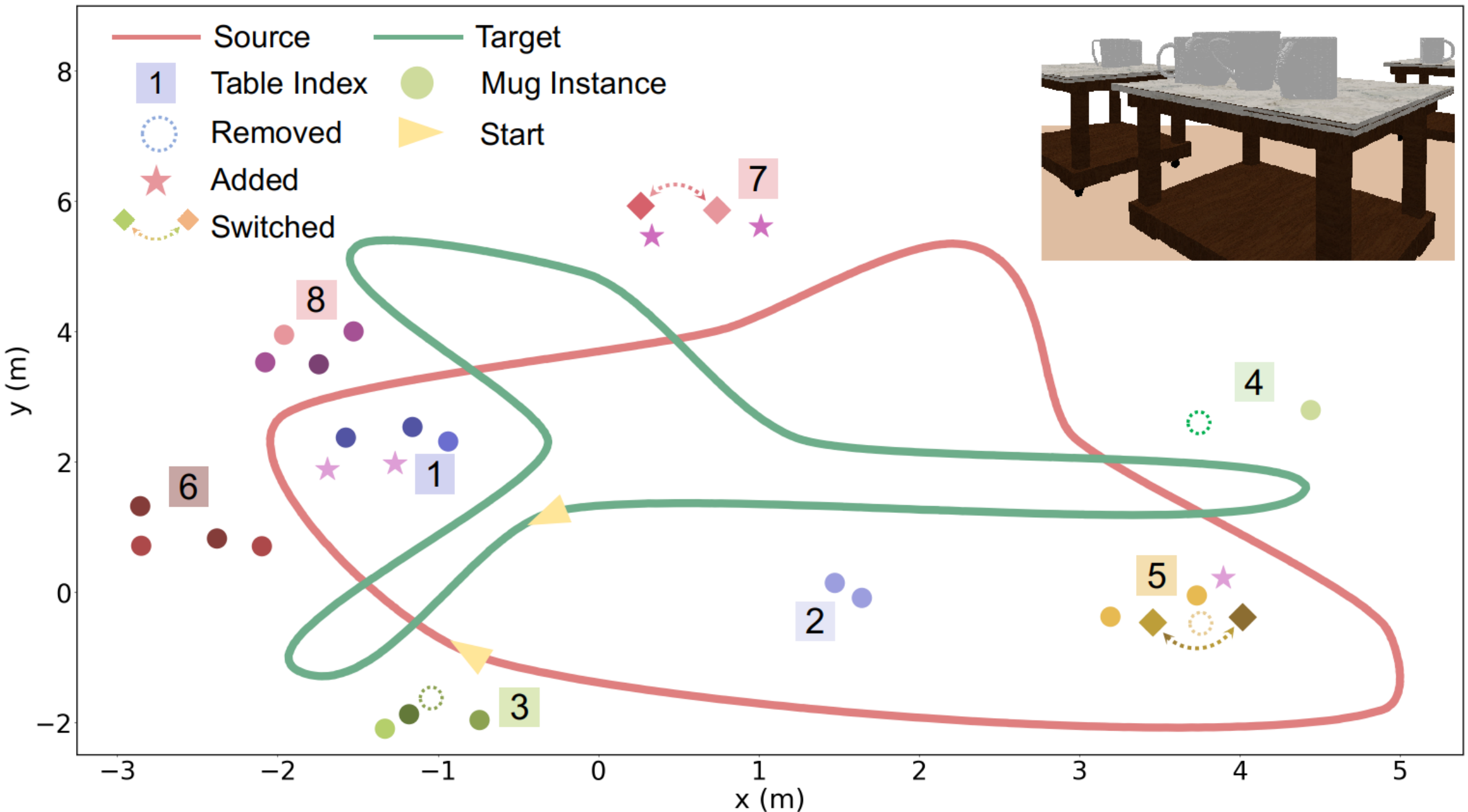}
    \vspace{-10pt}
    \caption{Synthetic scene layout and camera trajectories. Mug instance are shown in colored dots around eight tables as numbered squares. Changes of different types are shown with different icons. A snapshot of the rendered scene is shown in the upper right corner.}
    \label{syn_data}
    \vspace{-15pt}
\end{figure}
A local object graph $G_o = (V,E)$ for object $o$ in neighborhood $\mathcal{N}_0$ is constructed as a directed graph with vertices as all the objects in $\mathcal{N}_0$, $V=\{o_i\vert o_i \in \mathcal{N}_0\}$, and directed edges as the object center difference pointing from $o$ to its neighbors, $E=\{{\bf e}_{oi} \vert {\bf e}_{oi}={\bf t}_{i}-{\bf t}_{o},\forall i \in \mathcal{N}_0, i\neq o\}$. Considering the limited number of objects within a neighborhood (size set similar to a tabletop), the relatively small graph size makes it possible for edge-wise matching. We find corresponding edges ${\bf e}_{oj}$ and ${\bf e}_{o'j'}$ through vertex shape matching. Similar edges, indicating unchanged inter-object layout, is determined by the Euclidean distance between them:
\begin{equation}\label{edge}
\begin{aligned}
   \sum\limits_{j}\mathds{1}(\vert \vert {\bf e}_{oj}-{\bf e}_{o'j'} \vert \vert_2 \leq \delta_e) 
   = \begin{cases}0, & \text{changed}\\\geq 1, & \text{unchanged}.
\end{cases}
\end{aligned}
\end{equation}
If at least one pair of edges is found to be closer than $\delta_e$, i.e., at least one out of the few neighboring objects remains the same, the local object layout is marked as consistent. This implies an unchanged object, and vice versa. Note that in the rarer case when either $O$ or $o'_i$ is the only object in the neighborhood, we revert back to the object-wise comparison in terms of spatial proximity and shape similarity.
\begin{figure}
    \centering
    \includegraphics[width=0.90\linewidth]{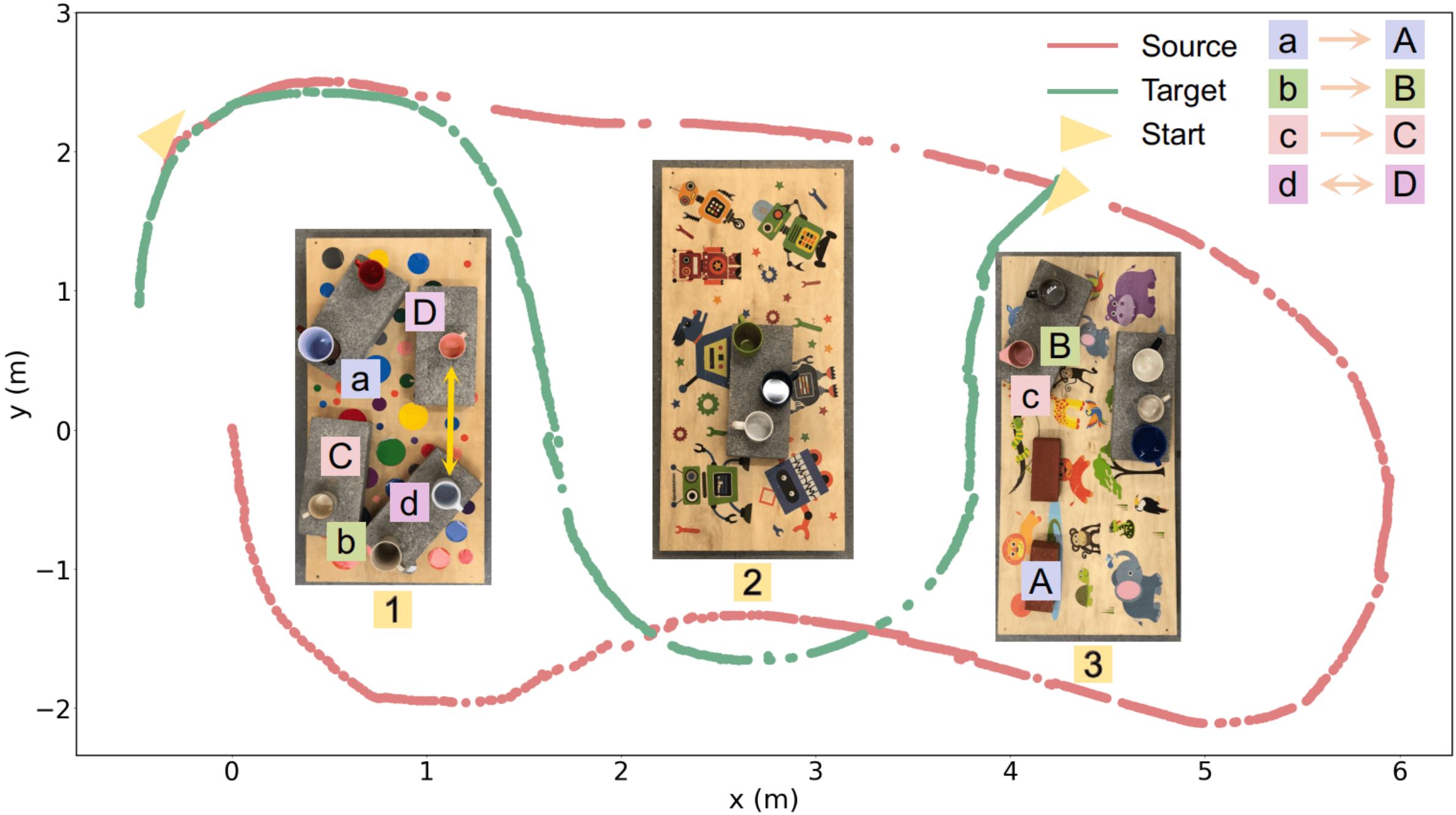}
    \vspace{-10pt}
    \caption{Real-world scene layout and camera trajectories. Changes take place on table 1 and 3, and are indicted by the colored letter block highlighted in the top view pictures of the three tables. }
    \label{real_data}
    \vspace{-10pt}
\end{figure}

Lastly, for detecting objects removed from the source, during target streaming we label each source object as \emph{observed} if it ever participates in shape matching with any target objects, then \emph{removed} objects can be found after the target sequence finishes as those \emph{observed} but never matched with a target object.

\section{Experiments and Results}
Targeting robust online change detection in the face of little observation overlap and localization errors, we would like to answer two questions: (1) Can we use the shape-consistent NDF-based representation as a valid category-level object representation that robustly generalizes to unseen object instances under partial observations? (2) Can our change detection approach perform well on sequences with little observation overlap and demonstrate robustness to localization errors?  We evaluate our approach on both synthetic and real-world sequences consisting of various mug instances, where mugs are added, removed, and switched places between sequences.

\subsection{Datasets}
To better examine the effectiveness of our \emph{category-level} object representation and the change detection approach built on top of it, we choose to have our testing sequences composed of objects from the same category. We hence design two pairs of testing sequences, in simulation and in real world, respectively, featuring coffee mugs of diverse shapes observed from distinct viewing angles. Considering the extensive multi-category  results reported in relevant works~\cite{9812146,mescheder2019occupancy}, we argue that the effectiveness of our approach should be extendable to the multi-category case by incorporating more object categories into the training data.
\begin{figure*}
    \centering
    \includegraphics[width=0.85\linewidth]{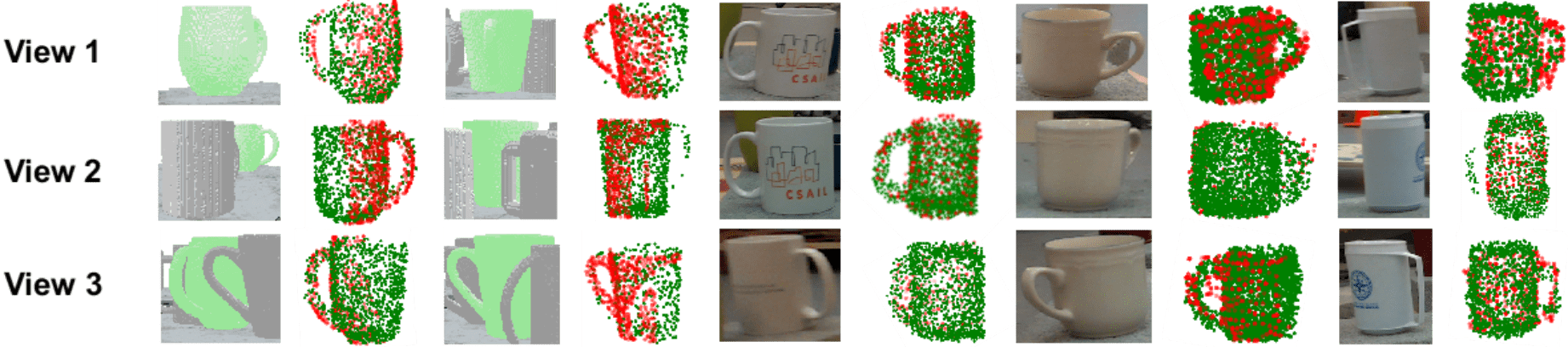}
    \vspace{-5pt}
    \caption{3-view  shape  reconstruction of  five  unseen  mugs (each column)  from  the synthetic  and  real-world  testing  sequences. For each view, Left: Partial observation. Right: Partial point clouds (red) and full shape reconstruction (green). For the first two columns of synthetic mugs, the two target mugs for reconstruction is highlighted in green. The point clouds are rendered in an orientation that showcases the boundary between partial observation and the predicted shape completion, whose shape fitness between the two colored point clouds demonstrates NDFs'  effectiveness in completing partial shapes.}
    \label{real_rec}
    \vspace{-15pt}
\end{figure*}

\textbf{Synthetic Sequences. }We set up an environment in Pybullet with 35-40 instances of 20 ShapeNet~\cite{chang2015shapenet} mugs models scattered around eight tables and obtain RGB-D images along with segmentation masks from two preset camera trajectories. The mug models are unseen by NDFs  during training. Each table supports 3-5 mugs with mugs newly added, removed, or switched locations, creating in total 12 changes between the two sessions. The two camera trajectories are designed such that the camera always faces towards the nearest table, prompting less observation overlap between sequences around table 1,2, and 5 (see Fig.~\ref{syn_data}).

\textbf{Real-world Sequences. }We collect two sequences featuring 14 mugs of diverse shapes randomly placed on three rectangular tables. We mount an Intel RealSense L515 camera onto a Jackal robot close to  mugs' height and collect RGB-D data with the trajectories shown in Fig.~\ref{real_data}. The camera trajectories are recovered using ORB-SLAM3~\cite{campos2021orb}. Since the camera is mounted facing sideways, while the camera circles around table 1, 2, 3 in the source, it only weaves through table 1 and 3 in the target, thereby creating observation disparity on mugs along the inner side of table 1 and 3.

\subsection{Metrics}
We adopt the commonly used precision and recall rate at the \emph{object} level, i.e., the number of objects,  for evaluation.  We report the number of correctly detected changes (true positives, TP), falsely detected changes (false positives, FP), and undetected changes (false negatives, FN). Precision and recall rate are computed as $\frac{TP}{TP+FP}$ and $\frac{TP}{TP+FN}$.

\subsection{Implementation Details}\label{implementation details}
To train NDFs using an occupancy network, we set the camera-table distances as $d_{c}=$ 0.2 m and $d_f=$ 5 m and generate 50,000 RGB-D partial observations with 94 ShapeNet~\cite{chang2015shapenet} mug models following the sample generation strategy in~\ref{gen_sim}. Partial object point clouds are obtained by extracting corresponding depth points in the camera frame indicated by the segmentation masks on RGB images. We build our model upon the network implementation provided by~\cite{9812146}
 and train it on two NVIDIA RTX 3090 GPUs using a learning rate of $6\times10^{-4}$ with the Adam optimizer.  The length of the interval tree  $l$ is set to be 1.2 m and 1.6 m for the synthetic and real-world sequence, respectively, intending to group most objects on the same table plane. For parameters in~\ref{change_detection}, we set $\delta_s=$ 0.9, $\delta_d =$0.02 m, and $\delta_e =$ 0.03 m. 

\subsection{Generalization to Unseen Instances}\label{transfer}

To demonstrate the robust generalization of the shape-consistent NDFs to unseen object instances under various viewing angles, in Fig.~\ref{real_rec}, we render the reconstructed 3D shapes of five mugs drawn from the synthetic and real-world testing sequences. The three views are selected in the hope of representing some of the most typical perspectives when observing a mug, e.g., handles visible in different directions and handle obscured. 
We can conclude that despite the variety of the selected mug shapes, the shape-consistent NDFs still demonstrate satisfactory performance in encoding and completing the full shape under varying viewing angles, justifying the adoption of NDF-derived object representation in our change detection task. 

This can be attributed to the fact that by learning purely from geometric structures embedded in the partial point clouds, NDFs are able to transfer this structural knowledge to unseen instances regardless of object color and texture. We also include cases with occlusion and shape ambiguity. For occluded observations, as shown in the first two columns of ``View 2" and ``View 3" , we can still see decent reconstruction results by virtue of the occlusion scenes incorporated in the training data. For the case of shape ambiguity, i.e., the handle is obscured in the second and last column of ``View 2", the major body parts of the mugs are still reconstructed reasonably well. While the handle location was incorrectly predicted, such mistakes only result in negligible errors to object center recovery.
\subsection{Results on Change Detection}
We further present the change detection results on the two sets of testing sequences based on our proposed approach and compare them to two baselines. 

The first baseline is the typical nearest neighboring point search (NN) commonly used for offline scene differencing~\cite{langer2020robust,ambrucs2014meta}. Given two spatially aligned observations $S$ and $T$, the change point set $C$ of $T$ w.r.t $S$ is found as $C = \{p\vert p\in T,  \forall s \in S, \vert \vert p-s \vert\vert_2>d\}$. We adapt it to the object-level and define that an object is \emph{changed} if a certain proportion $r$ of its target point cloud finds no neighbors in the source. To make the results interpretable to online object-level change detection evaluation, we mark an object as \emph{changed} the first time during data streaming  it is detected as \emph{changed}, as during real-world operation, corresponding actions will be taken right after changes are detected and usually no correction can be made for false positives.

The second baseline is the recently proposed panoptic multi-TSDFs (PMT) mapping method by Schimd et al.~\cite{schmid2022panoptic}, which represents panoptic entities as TSDF submaps and captures online object-level scene changes based on voxel-wise weight counting informed by TSDF value differences. After tuning the default parameters for better performance, we count the total number of changed objects based on the number of conflicting submaps determined by the baseline. 
\begin{figure}
    \centering
    \includegraphics[scale=0.2]{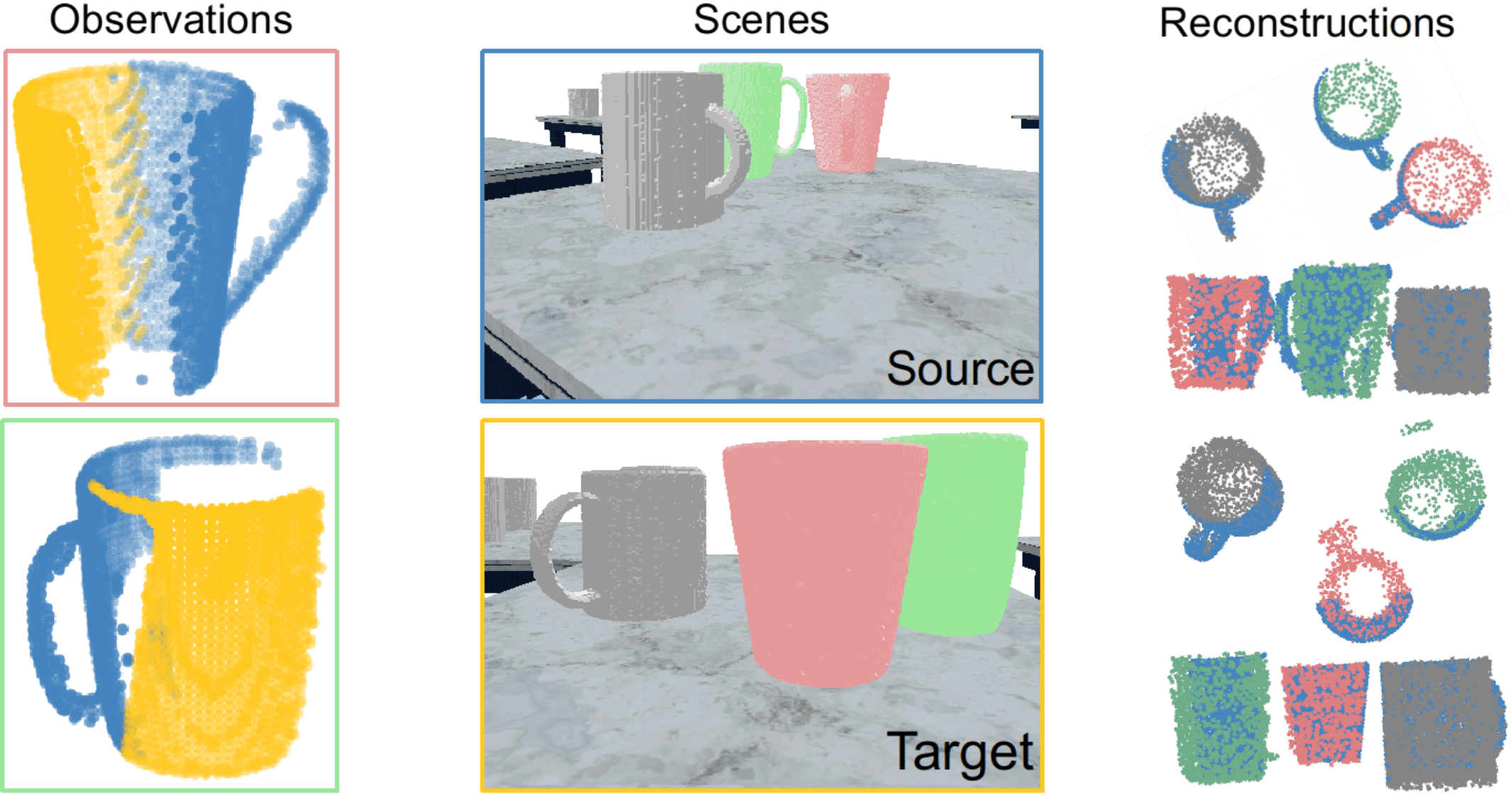}
    \vspace{-5pt}
    \caption{ The source  and target sequences share little observation overlap for the green and red mugs. Despite the distinct viewing angles and occlusion, (blue: from source, yellow: from target), our approach can still recover the shape and accurate object center from the NDFs representation, thereby successfully associating the target point clouds to the correct source mugs.}
    \vspace{-15pt}
    \label{syn_res}
\end{figure}
Since both baselines require pre-aligned sequences, we feed the baselines with ground truth camera poses from Pybullet and those aligned by ORB-SLAM3, respectively. For NN, the reference partial point cloud of each source object is fused from the whole sequence of the source depth images. Data associations between the target partial mug observations and the source point clouds are determined by the instance ID in Pybullet for synthetic sequences and manual labeling based on the panoptic masks produced by Detectron2~\cite{wu2019detectron2} (used in PMT) for real-world sequences.

All results are obtained on a laptop with an Intel Core i7-9750H CPU and an Nvidia GeForce RTX 2070 GPU. The network inference speed for NDF is around 0.019 s per frame, and the change detection operation with an expanding spatial tree executes at an average speed of 0.023 s per frame given the magnitude of the object number in our experiment setting (up to 40 mugs). These result in an overall speed of 24 FPS for the complete pipeline, thus showing no obvious latency for online operation.

\textbf{Results on Synthetic Sequences. } 
We present the change detection results of the three methods in Table.~\ref{syn}. Here for NN, we set $d=$ 0.002 m and $r=$ 0.3. We apply a 6 DoF random transform to the target camera poses when running our approach so as to simulate the unavailability of motion capture systems during common mobile robot operations.

We can see that our method accurately detects all the changes without any false positives. On the one hand, NN and PMT respectively ignore four and one changes for the two pairs of mugs switching positions at table 5 and 7, as the similar viewing angles of these mugs during the source and target traverse induce high overlap between the target and source point clouds, thereby confounding the baselines by the existence of neighboring points and similar local geometry.

On the other, both baselines falsely mark the four mugs on table 1 and 2 as \emph{changed}, which is due to the remarkable viewing angle difference on these four mugs between the source and target trajectory. As explained in Fig.~\ref{syn_res}, the two mugs highlighted in red and green are observed in almost opposite directions, which then leads to little/no overlap between the two partial point clouds (blue and yellow). Therefore, few neighboring points can be found between the source and target observations, while at the same time leads to drastic differences in the resulted TSDF values. In contrast, our method is capable of encoding the full mug shape  with a compact shape code from the partial observations and hence produces no false positives. 

\begin{figure*}
    \centering
    \includegraphics[width=0.9\linewidth]{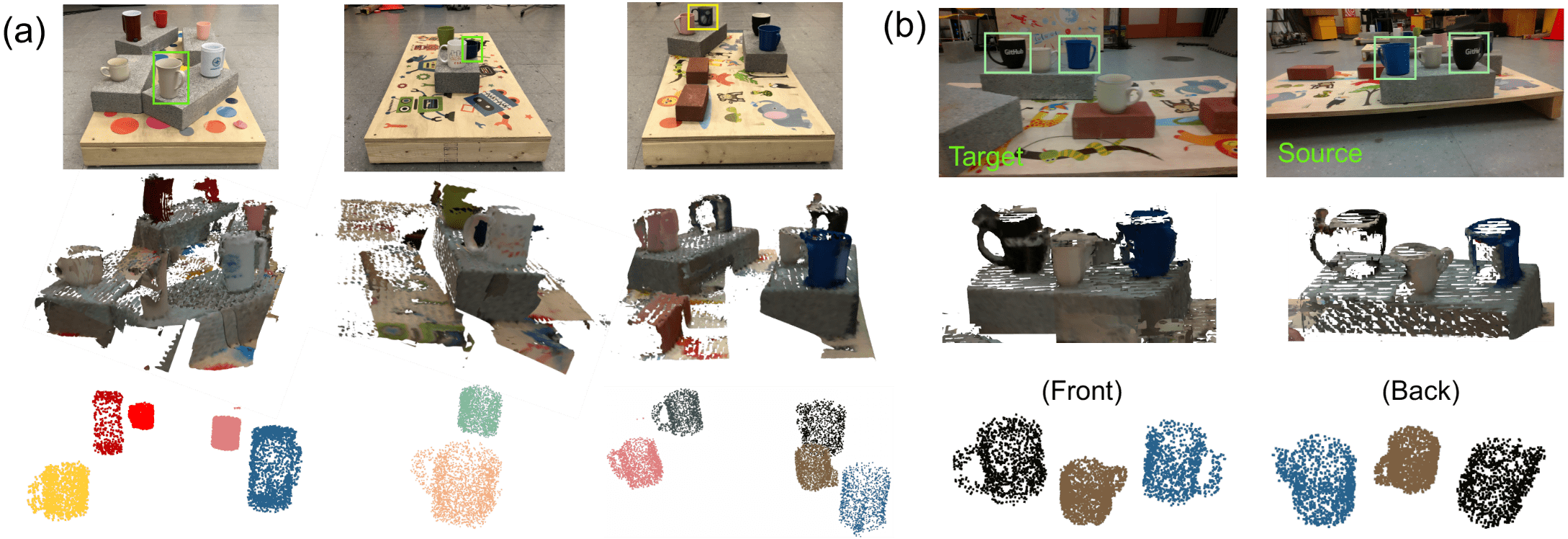}
    \vspace{-5pt}
    \caption{Top to bottom: real scenes, result from PMT, and results from NDFs. (a) Full reconstruction of the source real-world sequence. Front views of the three tables (first row) are included for mug layout illustration. Fully trained in simulation, NDFs are able to reconstruct 12 out of the 14 unseen mugs given partial observations, and the two missing mugs are highlighted in green in the front view pictures. (b) PMT and NDFs reconstruction results for the two mugs falsely detected as changes. As shown in the first row, two mugs (highlighted in green) are observed in almost opposite views in the source and target sequence. As shown in the second row, the missing part in the back of the PMT reconstruction misleads PMT to mark these two mugs as changed, while our NDF-based approach correctly recognize the mugs by virtue of the robust recovery of the full mug shape.}
    \label{real_res}
    \vspace{-15pt}
\end{figure*}

\textbf{Results on Real-world Sequences. }We report change detection results of the three methods on the real sequence in Table.~\ref{real} and present the reconstructions obtained from PMT and our method (for mugs) in Fig.~\ref{real_res}. For fair comparison, we obtain the mug point clouds for all the three methods using the panoptic mask adopted in PMT, which are generated by Detectron2~\cite{wu2019detectron2} and preprocessed by DBSCAN~\cite{ester1996density}. Here, the parameters of NN are set as $d=$ 0.002 m and $r=$ 0.4 considering the noises in the real-world point clouds. Due to localization drift between two sessions, for PMT, the final number of \emph{changed} objects are counted after manual merging of the few submaps instantiated for the same object. In order to mimic the real operating scenarios, the camera poses to our approach are obtained through running ORB-SLAM3 separately on the two traverses.

\begin{table}[]
\centering
\caption{Change Detection Results on the Synthetic Sequences. Best results are marked in bold.}
\vspace{-5pt}
\label{syn}
\begin{tabular}{|c|c|c|c|c|c|}
\hline
{ Approach} &
  { TP} &
  { FP} &
  { FN} &
  { Precision} &
  { Recall} \\ \hline
{ NN} &
  { 8} &
  { 4} &
  { 4} &
  { 66.7\%} &
  { 66.7\%} \\ \hline
{ PMT} &
  { 11} &
  { 4} &
  { 1} &
  { 73.3\%} &
  { 91.7\%} \\ \hline
{ Ours} &
  { \textbf{12}} &
  { \textbf{0}} &
  { \textbf{0}} &
  { \textbf{100\%}} &
  { \textbf{100\%}} \\ \hline
\end{tabular}
\vspace{-5pt}
\end{table}

\begin{table}[]
\centering
\caption{Change Detection Results on the Real-world Sequences. Best results are marked in bold.}
\vspace{-5pt}
\label{real}
\begin{tabular}{|c|c|c|c|c|c|}
\hline
{ Approach} &
  { TP} &
  { FP} &
  { FN} &
  { Precision} &
  { Recall} \\ \hline
{ NN} &
  { 3} &
  { 3} &
  { 2} &
  { 50\%} &
  { 60\%} \\ \hline
{ PMT} &
  { \textbf{5}} &
  { 3} &
  { \textbf{0}} &
  { 62.5\%} &
  { \textbf{100\%}} \\ \hline
{ Ours} &
  { \textbf{5}} &
  { \textbf{1}} &
  { \textbf{0}} &
  { \textbf{83.3\%}} &
  { \textbf{100\%}} \\ \hline
\end{tabular}
\vspace{-15pt}
\end{table}

When confronted with noisy point cloud inputs and localization results, despite the reconstruction failure of the beige mug on table 1 and the blue mug on table 2 (as shown Fig.~\ref{real_res}(a)) due to poor point cloud and panoptic mask quality, our method still maintains better detection efficacy in terms of both precision and recall rate.  From Fig.~\ref{real_res}(b), we see that the confusion brought by disparate viewing angles persists, as both NN and PMT wrongly mark the black and blue mug on table 3 (highlighted in green) as \emph{changed}. This matches the PMT reconstruction results shown in the second column, as the two mugs have just one side partially reconstructed during each session. Furthermore, another false detection emerges as the dark blue mug on table 3 (highlighted in yellow in Fig.~\ref{real_res}(a)), which results from the larger localization error at the starting point of the target trajectory. Our method reduces its reliance on the absolute accuracy of camera pose estimation by conducting local neighborhood layout matching, i.e., the relative spatial relationship among neighboring mugs, thereby successfully recognizing the mug through its unchanged relative 
positions with its neighbors.

\textbf{Limitation and Extension.} The proposed approach works well under the assumption that the changed objects do not take up a predominant portion of all the objects in the scene, which  is a common case for doing frequent visits of the same environment whose changes occur incrementally. For scenes with drastic changes between the two scans, our approach may not work well with the absence of a consistent local layout to refer to. Our approach can be further extended to detect objects that only rotate but do not move by utilizing the {\sothree} equivariance of NDFs, i.e., computing an extra {\sothree} transform between the two $\bf{z}$'s after the shape-similarity-based registration step. This situation was not included in the experiment as it is rare for changed daily objects to have pure rotation but no translation.

\section{Conclusion and Future Work}
In this paper, we propose an online object-level change detection approach based on an NDF-derived object representation, demonstrating improved robustness to viewing angle disparity and localization drift.  For future work, we would like to test the approach's scalability to larger scenes if incorporated as part of an object-level SLAM system targeting long-term operation, and further explore the potential of using NDFs for providing object pose constraints to help improve camera localization.

\renewcommand*{\bibfont}{\footnotesize}
\begin{flushright}
\printbibliography 
\end{flushright}

\end{document}